\documentclass[12pt]{iopart}
\usepackage{graphicx,import}
\expandafter\let\csname equation*\endcsname\relax

\expandafter\let\csname endequation*\endcsname\relax
\usepackage{amsmath}
\usepackage{amssymb}
\usepackage{floatrow} 
\usepackage{subfig}
\usepackage{lipsum}
\usepackage{anyfontsize}
\usepackage{mathtools}
\usepackage{cuted}
\usepackage[export]{adjustbox}
\usepackage{bibentry}
\usepackage{ulem}
\usepackage{xcolor}

\newcommand{\norm}[1]{\left\lVert#1\right\rVert}
\begin{document}

\title[Insight into Delay Based Reservoir Computing via Eigenvalue Analysis]{Insight into Delay Based Reservoir Computing via Eigenvalue Analysis}

\author{Felix Köster, Serhiy Yanchuk and Kathy Lüdge}

\address{Institut
for Theoretical Physics, Technische Universität Berlin, Berlin,
10559 Germany}
\ead{f.koester@tu-berlin.de}
\vspace{10pt}
\begin{indented}
\item[]Januar 2021
\end{indented}

\begin{abstract}
In this paper we give a profound insight into the computation capability of delay based reservoir computing via an eigenvalue analysis.
We concentrate on the task-independent memory capacity to quantify the reservoir performance and compare these with the eigenvalue spectrum of the dynamical system.
We show that these two quantities are deeply connected, and thus the reservoir computing performance is predictable by analyzing the small signal response of the reservoir.
Our results suggest that any dynamical system used as a reservoir can be analyzed in this way.
We apply our method exemplarily to a photonic laser system with feedback and compare the numerically computed recall capabilities with the eigenvalue spectrum.
Optimal performance is found for a system with the eigenvalues having real parts close to zero and off-resonant imaginary parts.
\end{abstract}

%
%
%
%
%

\section{Introduction}
Reservoir computing is a novel approach for time-dependent tasks in machine learning.
First introduced by Jaeger \cite{JAE01} and inspired by the human brain \cite{MAA02}, it utilizes the inherent computational capabilities of dynamical systems.  Very recently the universal approximation property has also been shown for a wide range of reservoir computers, which solidifies the concept as a broad applicable scheme \cite{GON20}.

Hardware setups have shown the feasibility and wide range of realizations \cite{FER03, ANT16, DOC09}, while theoretical and numerical analysis show interesting advancements \cite{GAL18a, GAL19} and pinpoint to easily implementable realizations \cite{ROE18a,GOL20}.
Different applications have been demonstrated \cite{BAU15,KEU17,SCA16,ARG17, ARG18,AMI19,PAT18, PAT18a,CUN19, VAN14}.
Since speed is of essence in computation, optoelectronic \cite{LAR12,PAQ12} and optical setups \cite{BRU13a,VIN15,NGU17,ROE18a,ROE20, BUE18a, BUE17, NAK16} are frequently studied, which additionally come with the benefit of low energy consumption. 

A new and sophisticated approach to the reservoir computing scheme was introduced by Appeltant et al. in \cite{APP11}, where a single dynamical node under the influence of external feedback utilizes a time-multiplexed reservoir.
The spatially extended network structure of classical reservoirs is no longer needed with this scheme, which reduces the complexity in reservoir hardware in exchange for processing speed.
A schematic sketch is shown in Fig. \ref{fig:sketch}.
Realizations with a single delayed reservoir \cite{ORT17a,DIO18,BRU18a,CHE19c,HOU18,SUG20} give a first glimpse over the potential of this idea for, e.g., time-series predictions \cite{BUE17,KUR18}, equalization tasks on nonlinearly distorted signals \cite{ARG20}, and fast word recognition \cite{LAR17}.
A general analysis, introduced by Dambre et al. \cite{DAM12}, was also used to quantify the task-independent computational capabilities of semiconductor lasers \cite{HAR19}.
For an overview, we refer to \cite{BRU19,SAN17a,TAN18a}.

A lot of research was already invested in order to develop a deeper understanding of reservoir computing systems, however, effective measures that allow to predict the performance are still missing. In this paper we want to fill this gap by providing a scheme that allows to predict general trends of the performance using the eigenvalue spectra of the dynamical system (the reservoir) without input. 
As an example reservoir, we chose a laser that is subjected to optical self-feedback. We use the Lang-Kobayashi system, which is an established model for a semiconductor laser with delayed external feedback. 
We calculate the total memory capacity as well as the linear and nonlinear contributions using the method derived in \cite{DAM12} and compare the results with the computed eigenvalue spectrum of the system, where we discover a clear connection. In particular, a high linear memory capacity is found for systems, where a large number of eigenvalues are close to criticality (with small negative real parts) and non-resonant (with imaginary parts not-resonant to the input timescale).

The paper is structured as follows.
First, we give an overview of the methods used for calculating the memory capacity and the eigenvalue spectrum in Sec.~\ref{sec:methods}.
After that, we present our results and discuss the impact of the eigenvalues on the performance and different nonlinear recall contributions first for a reservoir formed by a solitary laser and then by a laser with external cavity.

 

\section{Methods}
\label{sec:methods}
The reservoir computing scheme employs the idea of a dynamical reservoir, which projects input information into a high dimensional phase space. The nonlinear response of the reservoir is then used by a linear readout to approximate a specific task depending on the input.
Often the reservoir consists of many nodes with relatively simple dynamics (for example, $\tanh$-function \cite{JAE01}) in which the input enters via a weighted matrix. 
Afterward, the response is read out and linearly combined to generate an output. 
The idea is to minimize the Euclidean distance between the generated output and the target.
This approach is particularly resourceful for time-dependent tasks, because the dynamical system which forms the reservoir acts as a memory kernel.

\begin{figure}
	\centering
	\def\svgwidth{\textwidth}
	\import{}{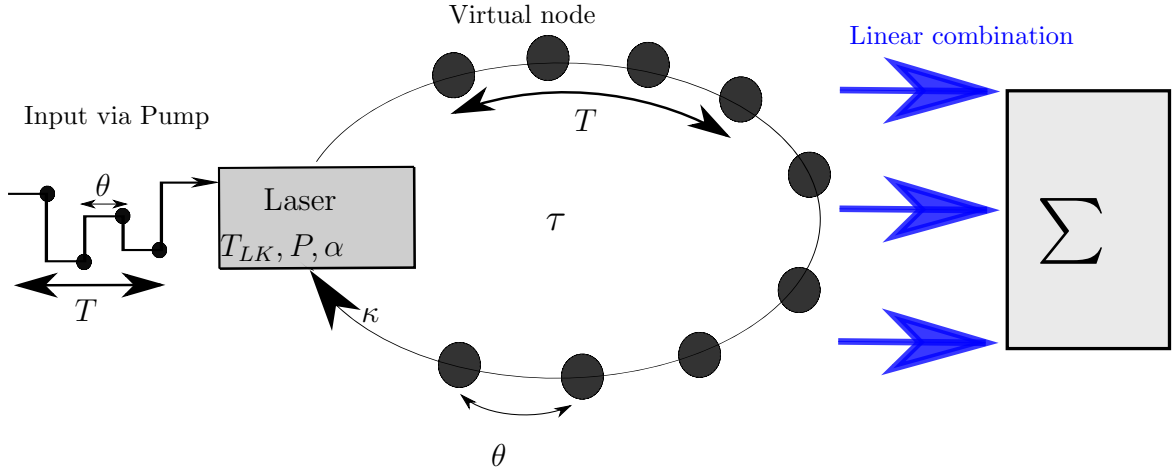}
	\caption[schematic]{Scheme of time-multiplexed reservoir computing with a laser. 
	}
	\label{fig:sketch}
\end{figure}

The modified approach introduced by \cite{APP11} uses a single node with delay as a reservoir, in which the output dimensions are distributed over time.
A mask $g$ is used to vary the input-signal in order to produce a high dimensional response. 
These responses are saved over time and used for the linear readout approximation. 
A sketch of the setup is shown in Fig. \ref{fig:sketch}.
In the following, we will give a short overview of the quantities and notations used in this paper.
We also refer to our previous works \cite{KOE20a,STE20}, where a detailed explanation of how the reservoir setup is operated and task-independent memory capacities are computed is given.

\subsection{Time-Multiplexed Reservoir Computing}
Let us briefly remind the main ingredients of the time-multiplexed reservoir computing scheme \cite{APP11,KOE20a,STE20}. An input vector $\mathbf{u}\in \mathbb{R}^L$ enters the system componentwise at times $t_l=lT$, $l=1,\dots,L$, $L$ being the number of sample points.
The time between two inputs $t_{l+1} - t_l$ is called the clock cycle $T$ and describes the period length in which one input $u_l$ is applied to the system.
Inside each interval of one clock cycle $T$, a $T$-periodic mask function $g$ is applied on the inputs (see Fig. \ref{fig:sketch}). The mask $g$ is piecewise-constant on $N_V$ intervals, each of length $\theta=T/N_V$ corresponding to $N_V$ virtual nodes.
The values of the mask function $g$ play the same role as the input weights in spatially extended reservoirs, with the difference that the input weights are now distributed over time.

The system responses are collected in the state matrix $\mathsf{S} \in \mathbb{R}^L \times \mathbb{R}^{N_V}$, where $N_V$ is the dimension of the measured system's state. 
More specifically, the elements of the state matrix are 
$\mathsf{S}_{ln}=s(lT+n\theta)$ with $n=1,\dots,N_V$, and $l=1,\dots,L$, where $s(t)\in\mathbb{R}$ is the state of the dynamical element of the reservoir at time $t$, e.g., a variable of the delay system in simulations, or laser intensity in an experimental realization.

A linear combination of the state matrix is given by $\mathsf{S} \mathbf{w}$, where $\mathbf{w}\in \mathbf{R}^M$  is a vector of weights. Such a combination is trained to find a least square approximation to some target vector $\mathbf{\hat{y}}$
$$
\arg\min_\mathbf{w} \left[ \| \mathsf{S} \mathbf{w} - \mathbf{\hat{y}} \|_2^2 + \lambda_{T} \| \mathbf{w} \|_2^2\right],
$$
where $\|\cdot\|_2$ is the Euclidean norm, and $ \lambda_{T}$ is a Tikhonov regularization parameter.  A solution to this problem is known to satisfy
\begin{align}
    \mathbf{w} = (\mathsf{S}^T \mathsf{S} + \lambda_{T} \mathrm{I})^{-1} \mathsf{S}^T \mathbf{\hat{y}},
\end{align}
when $\mathsf{S}^T \mathsf{S}+ \lambda_{T} \mathrm{I}$ is invertible. 
In the case of our Lang-Kobayashi laser model, since the physical system is intrinsically noisy, we used the state noise regularization \cite{JAE01,Jaeger2007} and set $\lambda_{T}=0$.
This is reasonable, as noise dominates very small dependencies in the given training data set, which the linear readouts would otherwise try to fit.
It also gives a more realistic threshold for the precision of the state readouts compared to numerical precision.
Comparisons of simulations without noise and with Tikhonov regularization to a noisy system without Tikhonov regularization yielded similiar results.

To quantify the system's performance, we use the normalized root mean square error (NRMSE) between the approximation $\mathbf{y}=\mathsf{S} \mathbf{w}$ and the target $\mathbf{\hat{y}}$
\begin{align}
   \text{NRMSE} = \sqrt{\frac{\sum\limits_{l=1}^{L}(\hat{y}_{l} - y_{l})^2}{N \cdot \mathrm{var}(\mathbf{\hat{y}})}} ,
\end{align}
where $\mathrm{var}(\mathbf{\hat{y}})$ is the variance of the target values $\mathbf{\hat{y}}=(\hat{y}_1,\ldots,\hat{y}_L)$.

\subsection{Memory Capacity}
Dambre et al. have shown in \cite{DAM12} that the computational capability of a reservoir system can be quantified via an orthonormal set of basis functions on a sequence of inputs. 
Here we give a recap of the used quantities introduced in \cite{KOE20a}.
In particular, the capacity to fulfill a certain task is given by
\begin{align} 
    \text{C}_{\mathbf{\hat{y}}} = \frac{\mathbf{\hat{y}}^T \mathsf{S} (\mathsf{S}^T\mathsf{S})^{-1} \mathsf{S}^T \mathbf{\hat{y}}}{\norm{\mathbf{\hat{y}}}^2}
    = 
    \frac{\mathbf{\hat{y}}^T \mathbf{y}}
    {\norm{\mathbf{\hat{y}}}^2}.
    \label{eq:mpsi_mem_capacity}
\end{align}
The capacity equals $1$ if $\mathbf{y}=\mathbf{\hat{y}}$ and the reservoir computer perfectly computes the task;
$C=0$ if it can not compute it at all, and inbetween $0$ and $1$ if it is partially capable to fulfill the task. 
In Sec. App. \ref{sec:APP}, we explain how Eq.~\eqref{eq:mpsi_mem_capacity} follows from the corresponding expression in \cite{DAM12}. 
Further, following Dambre et. al. \cite{DAM12}, we use finite products of normalized Legendre polynomials $\mathcal{P}_{d}$ as an orthogonal basis of the Hilbert space of all possible transformations (thus tasks with targets $\mathbf{\hat{y}}$) on an input sequence $\{u\} = \{ u_{-L}, \dots, u_{-3}, u_{-2}, u_{-1}\}$. As inputs into the system, we use uniformly distributed random numbers $u_l$, which are independent and identically drawn in $[-1,1]$. 
This yields uncorrelated inputs and thus uncorrelated memory capacities.
After feeding the input sequence $\{u\}$ of random numbers into the system, it yields a reservoir response $\mathsf{S}$. 
Formally, the memory capacity (Eq.~\eqref{eq:mpsi_mem_capacity}) is defined for an infinitely long sequence $L\to\infty$.
To approximate it numerically, we use $L=250000$.

In order to describe a task, the
target vector $\mathbf{\hat{y}}$ is defined as
\begin{align}
    \mathbf{\hat{y}}_{\{u\}} = \prod_{i} \mathcal{P}_{d_i}(u_{-i}),
    \label{eq:LP_construction}
\end{align}
where $\{d\} = \{d_1,...,d_I\}$ is a sequence of degrees such that the Legendre polynomial $P_{d_i}(u_{-i})$ of degree $d_i$ is applied to the input $u_{-i}$. The product of all such polynomials is used to generate the task (target vector $\mathbf{\hat{y}}$). 
The collection of all tasks \eqref{eq:LP_construction} for any possible degree sequence $\{ d \}$ is the Hilbert space of all possible transformations \cite{DAM12}. 

Further, to define the linear and nonlinear memory capacities, one uses special tasks, for which the sum of the degrees $\sum_i d_i$ is constant
\begin{align}
    \mathbf{\hat{y}}^d_{\{u\}} = \prod_{d_1+d_2+\cdots=d} \mathcal{P}_{d_i}(u_{-i}).
    \label{eq:tasks-d}
\end{align}
Clearly, there are many such possible tasks for all sequences $\{ d \}$ with $d=\sum_i d_i$. 
The memory capacity $MC^d$ of degree $d$ is defined as the sum of the capacities $\text{C}_\mathbf{\hat y}$ computed using Eq.~\eqref{eq:mpsi_mem_capacity} for all tasks \eqref{eq:tasks-d} of degree $d$: 
\begin{align}
    MC^d = \sum_{\{ d\}: \ d_1+d_2+\cdots=d}C_{\mathbf{\hat{y}}^d_{\{u\}}},
    \label{eq:memory_capacity}
\end{align}
The well known linear memory capacity corresponds to $d=1$.
The total memory capacity is then given by the memory capacities $MC^d$ of all degrees $d$. 
\begin{align}
    MC = \sum_{d>0} MC^d
    \label{eq:total_mem_cap}
\end{align}%
It was shown in \cite{DAM12} that $MC$ is limited by the readout-dimension $N_V$, which equals the number of virtual nodes $N_V$. 
An intuitive explanation is the following. 
The linear readout $\mathsf{S}\mathbf{w}$ of the reservoir computing scheme can be considered a linear combination of the columns of the state matrix $\mathsf{S}$.
Thus the amount of dimensions this basis can approximate is given by the number of linearly independent readouts. If the systems states are linearly independent, it can at most approximate $N_V$ different dimensions, which is in our case $N_V$ different tasks constructed from Eq. \eqref{eq:LP_construction}.
A more rigorous explanation is given by Dambre et. al. in \cite{DAM12}.

\subsection{NARMA10}
In addition to memory capacities, we evaluate the normalized root mean square error (NRMSE) of the NARMA10 task. 
NARMA10 is an often used benchmark test that combines linear and nonlinear memory transformations. It is given by the following iterative formula
\begin{align}
    A_{n+1} =  0.3A_n + 0.05A_n \left( \sum_{i=0}^{9}A_{n-i} \right) + 1.5 u_{n-9}u_n + 0.1.
\end{align}
Here, $A_n$ is an iteratively given number and $u_n$ is an independent and identically drawn uniformly distributed random number in $[0,0.5]$.
The reservoir is fed with the random numbers $u_n$ and has to predict the value of $A_{n+1}$.

\subsection{Lang-Kobayashi model}
We use the Lang-Kobayashi laser as an example reservoir.
This is a model applicable for semiconductor lasers with external feedback operating with low feedback strength.
The Lang-Kobayashi equations have been studied widely, modeling successfully semiconductor lasers \cite{ALS96,HEI99b} exhibiting complex dynamics and bifurcation scenarios \cite{ERN95a,ROT07,HEI99b}.
The dimensionless equations of motion are given by \cite{LAN80b}
\begin{align}
    \frac{dE(t)}{dt} &= (1 + i \alpha)N(t)E(t) + \kappa e^{i \phi}E(t - \tau) + D_{noise} \xi(t), \label{eq:LK_1}\\
    \frac{dN(t)}{dt} &= \frac{1}{T_{LK}}(P + \eta I(t)g(t) - N(t) - (2N(t) + 1)|E(t)|^2).
    \label{eq:LK_2}
\end{align}
The parameters scaling was chosen as in \cite{YAN10} with a modification to allow for the information input. The system time is normalized to the photon lifetime.
Here, $E$ is the complex electric field, $N$ is the charge carrier inversion, and $\xi$ describes spontaneous emission modeled by Gaussian white noise, $g$ is the masking function, $I$ is the input, $\alpha$ the amplitude-phase coupling, $\kappa$ is the feedback strength, $\phi$ the feedback phase, $\tau$ the delay time, $D_{noise}$ the noise amplitude and $P + \eta I(t)g(t)$ is the pump current, composed of $P$ a constant pump level and $\eta$ the input strength of the information fed into the system via eletric injection, which is small with respect to $P$.
$I(t)g(t)$ is the piecewise constant input function, which contains the data set values multiplied with a mask function.
$T_{LK}$ is the time scale ratio, modeling class B laser behaviour for sufficiently large $T_{LK}$, while $T_{LK} \ll 1$ models class A behaviour. $T_{LK} \approx 1$ are typical values in quantum dot \cite{ERN07a, OBR04,LIN15a} and quantum cascade \cite{WAN13c, COL14b} lasers while strong Class A lasers with $T_{LK} \ll 1$ are found in gas-laser systems. Note that the threshold pump current for the solitary laser is at $P_{th}=0$, while $P_{th}$ changes with $\kappa$ according to $P_{th} = -\kappa$.

\subsection{Calculating Eigenvalue Spectrum}
\label{sec:eig}
The eigenvalues of any dynamical system describe the dynamics for small perturbations around a linearized point. Because reservoir computers are often operated close to a stable equilibrium the eigenvalue spectrum of its linearized system can be analyzed.
The goal of this paper is to find a relation between the nonlinear memory recallability and the eigenvalue spectrum. The latter can be computed with much less numerical effort and could then be used to predict good parameter ranges for reservoir setups.
It also gives insight into the timescales of the eigendirections of the system, which contain information on the memory kernel of the reservoir.
To compute the eigenvalue spectrum, we used two methods:
the first method is an analytical approximation in the long delay limit \cite{LIC11}, while the second relies on numerical computation with the DDE-biftool software package \cite{ENG02,SIE14a,JAN10}.

To begin with, we give a short overview of the first method from \cite{LIC11}, which provides an approximation of the spectrum of long-delay systems. In \cite{YAN10}, it was applied to the Lang-Kobayashi system. As delay-based reservoir computing is mostly used with a long delay compared to the local dynamics, this is a valid approximation that gives a general tool to analyze the reservoirs of such type. 

The characteristic equation for the eigenvalues is obtained through the linearization around a steady state $x^*$, and it reads as
\begin{align}
    \det\left(-\lambda I + B + Ce^{-\lambda \tau}\right) = 0
    \label{eq:char_eq}
\end{align}
with some constant matrices $B$ and $C$, and $I$ is the identity matrix.
For large $\tau$, its solutions can be decomposed into two parts, in which one scales as $\Re(\lambda) \sim 1/\tau$, also called the pseudocontinous spectrum, and a strongly unstable spectrum with the scaling $\Re(\lambda) \sim 1$ with $\Re(\lambda)>0$. The strongly unstable spectrum is absent for reservoir computing applications, since, otherwise, the reservoir's 
state is strongly unstable, and the echo state (or fading memory) property \cite{JAE01} is lost to a large extend.
Formally, the condition for the absence of the strong unstable spectrum is the stability (all eigenvalues have negative real parts) of the linearization matrix A of instantaneous terms (see Eq. 11).
Hence, we focus on the pseudocontinous spectrum, which can be obtained by introducing the Ansatz
\begin{align}
    \lambda = \frac{\gamma}{\tau} + i \mu,
    \label{eq:subst_yan_method}
\end{align}
where $\gamma$ and $\mu$ are two new real variables.
Subsituting Eq. (\ref{eq:subst_yan_method}) into (\ref{eq:char_eq}) one gets in the leading order
\begin{align}
\label{eq:PCS}
    \det \left( -i \mu I + B + Ce^{-\gamma -i\mu\tau}\right) = 0.
\end{align}
Equation (\ref{eq:PCS}) is a polynomial with respect to  $e^{-\gamma -i\mu\tau}$. If $Y_j(\mu)$ are solutions of this polynomial, then
\begin{equation}
    \label{Y}
    e^{-\gamma -i\mu\tau} = Y_j(\mu), 
\end{equation}
and $\gamma_j(\mu)=-\ln |Y_j(\mu)|$ are the rescaled 
real parts of the eigenvalues from the pseudocontinuous spectrum. 
More exactly, the curves $\gamma_j(\mu)/\tau + i \mu$ are approximated with the eigenvalues for large $\tau$.

In the case of the Lang-Kobayashi system, there are two solutions for the real parts of the pseudocontinous eigenvalue spectrum, see the derivation in \cite{YAN10}, 
\begin{align}
    \gamma_1(\mu) &= -\ln\left|1 + i \frac{\mu}{\kappa}\right|, \label{eq:pcs1} \\
    \gamma_2(\mu) &= -\ln\left|1 + i \frac{\mu}{\kappa} + \frac{2\epsilon (P + \kappa)(\epsilon(1+2A^2) - i\mu}{\kappa(\mu^2 + \epsilon^2(1+2A^2)^2)}\right| ,
    \label{eq:pcs2}
\end{align}
where $\epsilon =  T_{LK}^{-1}$, and $A^2=\frac{P - N^*}{2N^* + 1}$ is the constant intensity, with the corresponding inversion $N^*$ at the external cavity mode (ECM). ECMs are the solutions of the Lang-Kobayashi system of the form $E=Ae^{i\omega t},N=N^*$ with constant $A$ and $N^*$, which play the role of the equilibria. 
Due to the $S^1$ symmetry of the system, each of these solutions can be transformed into an equilibrium with the corresponding characteristic equation (\ref{eq:char_eq}). 
The Lang-Kobayahi system possesses many ECMs, however, for the case $\alpha=0$, we consider the ECM with
$N^*=-\kappa$, which is the most stable \cite{YAN10}.
In general our method of linearization applies to all equilibrium solutions of any system, thus $\alpha \neq 0$ can be analyzed as long as the system is in equilibrium. For $\alpha \neq 0$ the system often starts to jump between different solutions, where some are not of equilibrium nature. 
Thus comparisons between the different operating points is not as simple and the focus on the essential new method gets lost.

In order to  approximate the imaginary parts $\mu$ of the pseudocontinuous spectrum, we consider the argument of \eqref{Y} and obtain 
\begin{equation}
\label{eq:muk}
    \mu_{j,k}  = \frac{2\pi k}{\tau}  -\frac{1}{\tau} \arg Y_j(\mu_{j,k}) ,\quad k\in \mathbb{N}
\end{equation}
where $\mu_{j,k}$ is the imaginary part of a $k$-th eigenvalue on the $j$-th branch. For the purpose of this paper, we need an approximation of the eigenvalues around the origin. As one can simply show using \eqref{eq:muk}, for large $\tau$, these eigenvalues (their imaginary parts) can be approximated as 
\begin{equation}
   \mu_{j,k}  = \frac{2\pi k}{\tau}  -\frac{1}{\tau} \arg Y_j(0) +\mathcal{O}\left(\frac{1}{\tau^2}\right),
\end{equation}
as soon as $k/\tau \ll 1$, see also \cite{YAN05, YAN14a} for more detailed estimations. 
In the case of the Lang-Kobayashi system, the roots $Y_j(0)$ are real, hence, we have either $\arg Y_j(0)=0$ or $\pi$. This leads to 
\begin{equation}
 \mu_{j,k}  \approx \frac{\pi }{\tau}(2k-\nu)  
\end{equation}
with $\nu=0$ for $\arg Y_j(0)=0$, and $\nu=1$ for $\arg Y_j(0)=\pi$. Hence, all imaginary parts $\mu_{j,k}$
are integer multiples of $\pi/\tau$. In particular, for any $T\approx j\tau$, $j\in \mathbb{N}$ ($T\approx j\tau/2$ for $\nu=0$, respectively), the product $\mu_{j,k} T$ is proportional to an integer number of $\pi$. This kind of resonance occurs for all considered eigenvalues (independent of $k$), and it plays an important role in the total memory loss of the reservoir, which is discussed in Sec.~\ref{sec:LK} below.

The second method for computing the eigenvalue spectrum is based on DDE-biftool \cite{ENG02,SIE14a}, which is a path-continuation package for Matlab capable of numerically computing the eigenvalues. 
In our case, we compute the first 100 eigenvalues with the highest real parts to compare these with results from the memory capacities. 

We also consider the case of no feedback $\kappa=0$. This yields a solitary semiconductor laser system that can be tuned from being an effectively 1-dimensonal problem (Class A-like for $T_{LK} \ll 1$) to a 2-dimensional problem (Class B-like for $T_{LK} \gg 1$) \cite{ARE84}. 
We will use a linearization and numerical evaluation of the eigenvalue problem. 
Even though the laser system is 3-dimensional, it possesses the $S^1$ symmetry $E\to E e^{i\phi}$ allowing to reduce the dimension by one. \\
We would like to emphasize that the eigenvalue method would also apply to a reservoir computer with a different form of information input, e.g. optical injection in the case of the Lang-Kobayashi system, since the analysis is performed without any reservoir computer input. As long as the input is a small perturbation to the system, the responses are fully described by the linearized system.

\subsection{Simulation description}
Simulations have been performed in C$++$ with standard libraries, except for
linear algebra calculations, which were done via the linear algebra library "Armadillo" \cite{SAN16}.
A Runge-Kutta 4th order method was applied to integrate the delay-differential equation given by Eqs. (\ref{eq:LK_1}) and (\ref{eq:LK_2}) numerically, with an integration step $\Delta t=0.01$ in time units of the photon lifetime.
The noise strength is $D_{noise}=10^{-7}$ in all simulations.
After simulating the system without reservoir inputs to let transients decay, a buffer time of 100000 inputs was applied (this is excluded from the training process).
In the training process, 250000 inputs were used to have sufficient statistics ($N=250000$). Afterward, the memory capacities were calculated, whereby a testing phase is not necessary, because the independent and identically drawn uniformly distributed inputs $u$ are statistically equal if drawn for training or testing phases.
All possible combinations of the Legendre polynomials up to degree $D=5$ and 500 input steps into the past were considered ($i=-500$). Capacities $C_{\mathbf{\hat{y}_{\{ u\}}}}^d$ below $0.001$ were excluded because of finite statistics. For calculating the matrix inverse, the Moore–Penrose pseudoinverse from the C++ linear algebra library "Armadillo" was used.
In the case of the NARMA10 task, 25000 inputs for training and testing were used.
For the piecewise-constant $T$-periodic masking function $g$ independent and identically distributed random numbers between $[0,1]$ were used.

For all simulations, the input strength $\eta$ was fixed to $0.01$. The small input strength was used to guarantee the linear answers of the reservoir and, hence, the relevance of the eigenvalue analysis.
\subsection{Geometrical intuition}
\begin{figure}
\centering
  \def\svgwidth{\columnwidth}
    \resizebox{\textwidth}{!}{\import{}{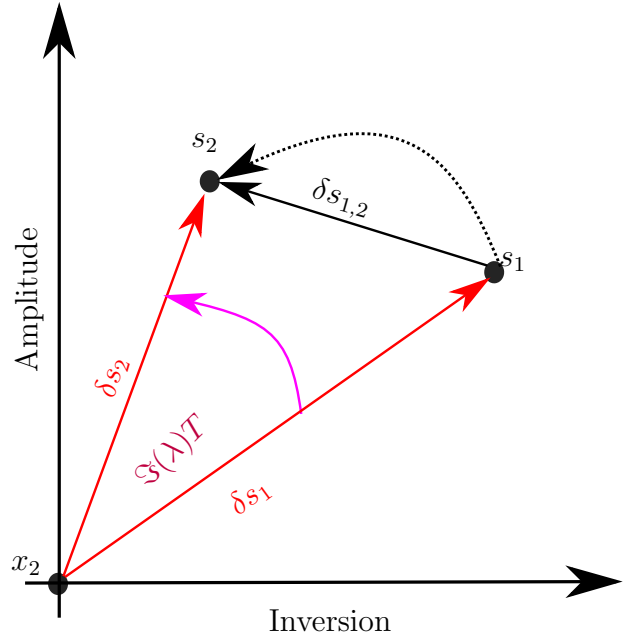}}
  	\caption[schematic]{Sketch of the system response in phase space to a small input during the clock cycle $T$. The trajectory moves from a state $s_1$ to a new state $s_2$ (dotted black line). $x_2$ is a equilibrium of the system due to the new reservoir input $u_2$. The red vectors $\delta s_1=s_2 - x_2$, $\delta s_2=s_1-x_2$ indicate the distances  from this equilibrium for the two instances in time. The distance reduction $\Lambda=e^{\Re(\lambda)T}=|\delta s_2 |/|\delta s_1 |$ describes the relation of the magnitudes of the two vectors. The purple arrow describes the angular distance $\Phi = \Im(\lambda)T$ covered in one clock cycle interval $T$. For simplicity, we excluded trajectory responses for different virtual nodes $\theta$.	
  	}
  \label{fig:illustration}
	\centering
	\def\svgwidth{\textwidth}
\end{figure}

In this paper we will use two quantities $\Phi$ and $\Lambda$ to approximate the memory capacity properties of the reservoir computer. For these two values we would like to give a geometrical intuition, shown Fig. \ref{fig:illustration}.

The first value $\Phi=\Im(\lambda)T$ we call the relative angular distance between two inputs, where $\Im$ denotes the imaginary part of the eigenvalue $\lambda$.
Here $\lambda$ is a critical eigenvalue, i.e. one having its real part close to $0$.
$\Phi$ geometrically describes the angular distance between two distance vectors $\delta s_1$ and $\delta s_2$ of the system's state $s_1$ and $s_2$ at two instances in time separated by one clock cycle interval $T$.
If this relative angular distance is a multiple of $\pi$ the responses tend to overlap, reducing the separability of the inputs, thus degrading the reservoir computer performance.

The second quantity $\Lambda=e^{\Re(\lambda)T}=|\delta s_2 |/|\delta s_1 |$ describes the distance reduction between two perturbed states, where $\Re$ denotes the real part of the eigenvalue $\lambda$.
$\Lambda$ describes the contraction of the system's state towards a new equilibrium due to a new reservoir input.
To distinguish two responses $s_1$ and $s_2$ for two different inputs $u_1$ and $u_2$, the distance $|\delta s_{1,2}|$ (see Fig. \ref{fig:illustration}) between two responses should be large enough. 
On the other hand, if the reaction of the system is very fast, i.e. very negative eigenvalues, the system has a high echo state property and thus low memory capacities for any inputs longer than a few (in the worst case even longer than one) steps into the past. 
If the remaining information of the input n-th steps back is degrading very fast (very negative eigenvalues), the systems capability to recall is lowered, and at some point reaches the level of the system noise.
The distance reduction $\Lambda = e^{\Re(\lambda)T}$ gives a good estimation for both of these properties.

In this paper we will show that both quantities together pinpoint to well performing reservoir computer setups.
\begin{figure}%
    \centering
    \includegraphics[width=\textwidth]{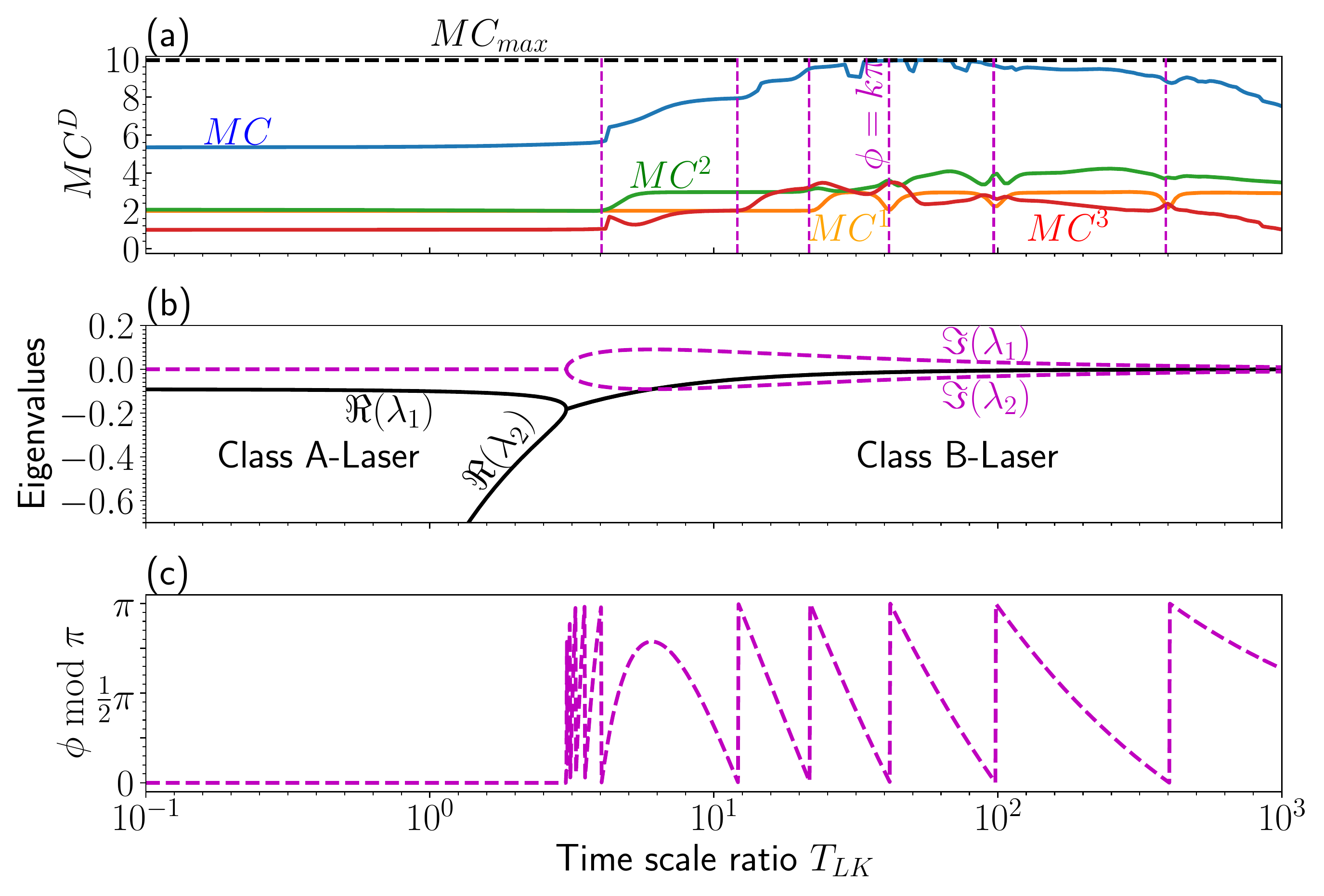}
    \caption{(a) Linear $MC^1$, quadratic $MC^2$, cubic $MC^3$, and total $MC$ memory capacities plotted over the logarithmic lifetime scale ratio $T_{LK}$. 
    (b) The real and imaginary parts of the eigenvalues showing the transition from class A to class B system. (c) The angular distance between two inputs $\Phi=\Im(\lambda)T$ taken modulo $\pi$. 
    Results shown are for the laser system with $\kappa=0$, $P=0.05$, $N_V=10$ and $T=220$ and a logarithmic scan for $T_{LK}$.
    }%
    \label{fig:class_B_eigenvalue_memory_cap}%
\end{figure}%

\section{Results}
\label{Results}
This section is structured as follows.
First, we will discuss a Lang-Kobayashi laser with $\kappa=0$, i.e. a solitary laser system, as a reservoir to simplify and depict general concepts. Afterward, we will activate the delay and look at the full Lang-Kobayashi system as a reservoir computer.
\subsection{Laser without feedback}
\label{sec:2d}
We first consider a solitary semiconductor laser system as a reservoir.
One has to think of the virtual nodes not to be located on the delay line, but rather as time separated readouts of the system state, that are used in a linear combination. 
We set $\kappa=0$ in Eq. (\ref{eq:LK_1})--(\ref{eq:LK_2}) and use 10 virtual nodes ($N_V=10)$. 
For the considered parameter values $P=0.05$ and $\alpha=0$ and without input and noise ($\eta=0$, $D_{noise}=0$), the system's solution converges to a single stable ECM, for which we compute the two eigenvalues. 
The two eigenvalues are plotted in Fig.~\ref{fig:class_B_eigenvalue_memory_cap}(b) as a function of $T_{LK}$, which gives from left to right the transition from Class A to Class B laser. 

To compare the two eigenvalues with the recall capability of the laser, we plot the computed linear, quadratic, cubic, and total memory capacities in Fig.~\ref{fig:class_B_eigenvalue_memory_cap}(a).
The memory capacities do not change significantly for $T_{LK}\lesssim3$ where the system corresponds to a class A laser with an adiabatic approximation of the charge carriers. For these parameter values, as one can see from the real parts of the eigenvalues, one eigendirection is considerably faster than the other, and thus can be ignored. 
At $T_{LK} \approx 2$, the transition from a Class A laser to a Class B laser appears, whose steady state solution is a focus. The additional degree of freedom of the charge carrier dynamics leads to an increase of the total memory capacity from about 5 to the theoretical maximum of 10.

Fig.~\ref{fig:class_B_eigenvalue_memory_cap}(c) shows the angular distance
$\Phi = \Im(\lambda)T$ taken modulo $\pi$, which 
is based on the rotation $\Im(\lambda)T$ of a small perturbation vector in the 2-dimensional phase space during the evolution over the time-interval $T$ (see Fig. \ref{fig:illustration}). 
The discontinuities of $\Phi$ in Fig.~\ref{fig:class_B_eigenvalue_memory_cap}(c) (indicated with vertical dashed purple lines in Fig.~\ref{fig:class_B_eigenvalue_memory_cap}(a)) correspond to resonances, i.e. integer numbers of half-a-circle rotations.
Comparing the memory capacity at these points in the class B regime, one observes dips in the linear memory capacity and slight changes in the higher-order memories.
This effect is pronounced if at the same time real parts of the eigenvalues are close to 0.
Since the degradation of the linear memory coincides with the discontinuities in $\Phi \mod \pi$, it can be linked to an overlapping of the systems responses, and to a decreasing linear separability of the output.
We would like to emphasize that even though the system has no optical feedback ($\kappa = 0$) the dynamical system still can act as a reservoir with very short memory. This comes from the fact that the reaction of the system is not instantaneous yielding a memory kernel of a few inputs into the past. 
The memory of the reservoir is limited by the real part of the largest critical eigenvalue.
This is due to the fact, that the real part of the largest eigenvalue yields the timescale on which small perturbations to the equilibrium exponentially decay.

\begin{figure}%
    \centering
    \includegraphics[width=\textwidth]{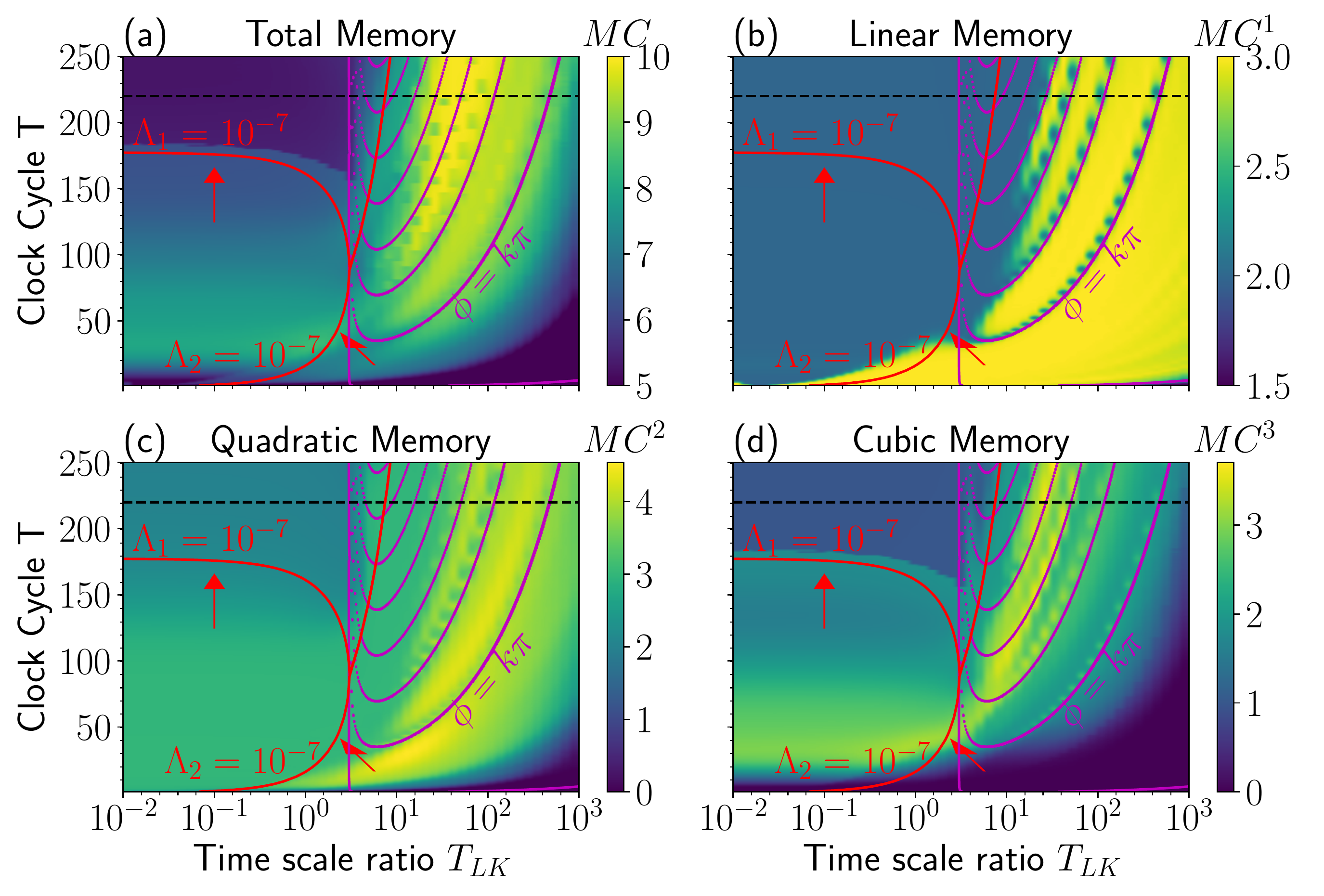}
    \caption{2-dimensional parameter scan in the plane of clock cycle $T$ and logarithmic life time scale ratio $T_{LK}$ showing the total, linear, quadratic, and cubic memory capacities as a color code in panel (a)-(d). The purple and two red solid lines show the parameter values where $ \Im(\lambda) T  \approx k \pi$, $k=1,2,\dots$, and $\Lambda_{1,2} = e^{\Re(\lambda_{1,2})T} \approx 10^{-7}$, respectively. $\lambda_{1,2}$ are the two eigenvalues of the laser system shown in Fig. \ref{fig:class_B_eigenvalue_memory_cap}. 
    The dashed black line indicates the parameter scan used in Fig. \ref{fig:class_B_eigenvalue_memory_cap}
    Other Parameters as in Fig. \ref{fig:class_B_eigenvalue_memory_cap}.}%
    \label{fig:class_B_2d_scan}%
\end{figure}

For a larger picture of the resonance effects at $\Phi \approx k\pi$ ($k \in \mathbb{N}$), a 2-dimensional parameter scan was done as a function of the timescale ratio $T_{LK}$ and the clock cycle $T$ (shown in Fig.~\ref{fig:class_B_2d_scan}).
The linear, quadratic, cubic, and total memory capacities are color-coded in panel (a-d).
Bright regions in (a-d) correspond to high memory capacities, while dark blue to low memory capacities.
The black dashed line shows the scan from Fig.~\ref{fig:class_B_eigenvalue_memory_cap}. 
Purple solid lines show the parameter values where $\Phi \approx k\pi$.
The influence of the angular distance $\Phi$ is most prominent in Fig.~\ref{fig:class_B_2d_scan}(b), where dips are visible in the linear memory capacity. Its influence on the higher-order memory capacity is also detectable, but harder to describe, as both quadratic and cubic memories either decrease or increase depending on the resonance line. \\
The solid red lines denote parameter values, where the distance reduction $\Lambda = e^{\Re(\lambda)T}= D_{noise} = 10^{-7}$ for the two eigenvalues $\lambda_{1,2}$ of the solitary laser system. 
The two red arrows indicate the direction in the parameter space for decreasing $\Re(\lambda)$ and thus decreasing $\Lambda$. 
The distance reduction shows a decrease in the memory capacities for a decrease in $\Lambda$.
This rises from the fact that lower $\Lambda$ correspond to faster eigendirections and thus faster echo state properties. \\
Combining the information about the two quantities $\Phi$ and $\Lambda$ and comparing it with the memory capacity, we can pinpoint to well performing reservoir computers for the class B and class A laser system.
Namely, the linear memory capacity has larger values in the absence of resonances $\Phi \approx k \pi$ and for values of $\Lambda$ closer to 1. We now want to expand this knowledge to the case of a laser with external feedback.

\subsection{Laser with feedback}
\label{sec:LK}
We now expand our results to the infinite-dimensional phase space of a laser system with delay, i.e., the Lang-Kobayashi system.
In \cite{PAQ12,ROE18a, ROE20, KOE20a,STE20} it was shown that resonances between $\tau$ and $T$ often decrease memory capacity and thus reservoir computing performance.
Here we look at this phenomenon from another point of view, namely, as a resonance between $T$ and the imaginary parts of the eigenvalues. 
We use the resonance property 
described in Sec.~\ref{sec:eig}: for certain resonant values
of $T$, the product $\Im(\lambda)T$ is proportional to an integer number of $\pi$ for all critical eigenvalues simultaneously.
We computed the first 100 eigenvalues using DDE-biftool for the Lang-Kobayashi system. By superimposing all $\Phi_i$, where $i \in 0,1,2...,N$ is the index of the $i$-th eigenvalue, we evaluate the resonance effects of the strongest $N$ eigendirections by computing the average angular distance
\begin{align}
\label{eq:hatPhi}
    \hat{\Phi} = \frac{1}{N}\sum_{i=0}^N \Phi_i,
\end{align}
and compare the results with the linear, quadratic, and total memory capacities. 
\begin{figure}%
    \centering
    \includegraphics[width=\textwidth]{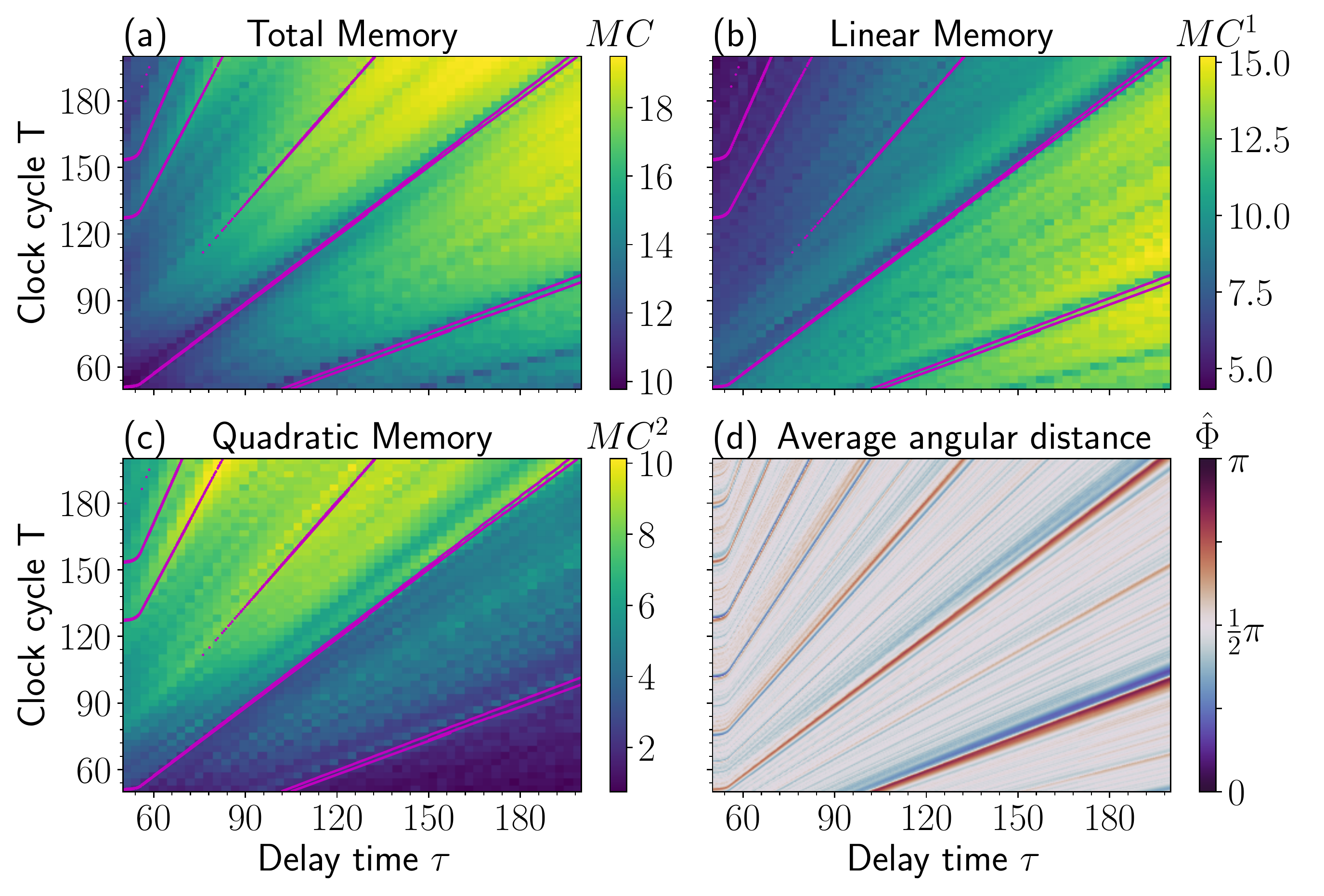}
    \caption{2-dimensional parameter scan in the plane of clock cycle $T$ and delay $\tau$ with $T_{LK}=100$, $N_V=20$, $P=0.01$, $\kappa=0.1$. Color coded is (a) total, (b) linear, and (c) quadratic memory capacities.(d): Average angular distance $\hat \Phi$ given by Eq. \eqref{eq:hatPhi} for the first 100 eigenvalues. Parameter values where 
    $\hat\Phi \approx k\pi$ are shown by solid purple lines in panels (a)-(c).
    }%
    \label{fig:2D-scan_with_delay_imaginary_superposition}%
\end{figure}%
The comparison of the memory capacities and $\hat\Phi$ is shown in Fig.~\ref{fig:2D-scan_with_delay_imaginary_superposition}, where
a 2-d parameter scan is plotted in the parameter plane of the delay time $\tau$ and the clock cycle $T$.
Bright regions in (a-c) correspond to high memory capacities, while dark blue to low memory capacities.
Panel (d) shows the results for $\hat{\Phi}$ for the first 100 eigenvalues.
Values close to $0$ or $\pi$ indicate parameters where
all leading eigendirections possess resonant eigenvalues, i.e., 
$\Im(\lambda)T\approx k\pi$, and perform an integer of half circle rotations during one input time $T$.
The solid purple lines in Fig. \ref{fig:2D-scan_with_delay_imaginary_superposition}(a-c) indicate the resonant values $\hat\Phi\approx k\pi$.
A match with lower total memory capacities, especially for the linear memory is clear.
For a reservoir to be applicable to many tasks, a higher total memory capacity is desirable.
Our results support the fact that the clock cycle $T$ should be chosen to be off resonant of the delay time $\tau$.
The eigenvalue analysis gives an additional explanation and intuition for why this is the case.
Taking into account the resonance effect and our results from \cite{KOE20a}, we set the delay time $\tau\approx \sqrt{2} T$ for all following simulations. 
\begin{figure}%
    \centering
    \includegraphics[width=\textwidth]{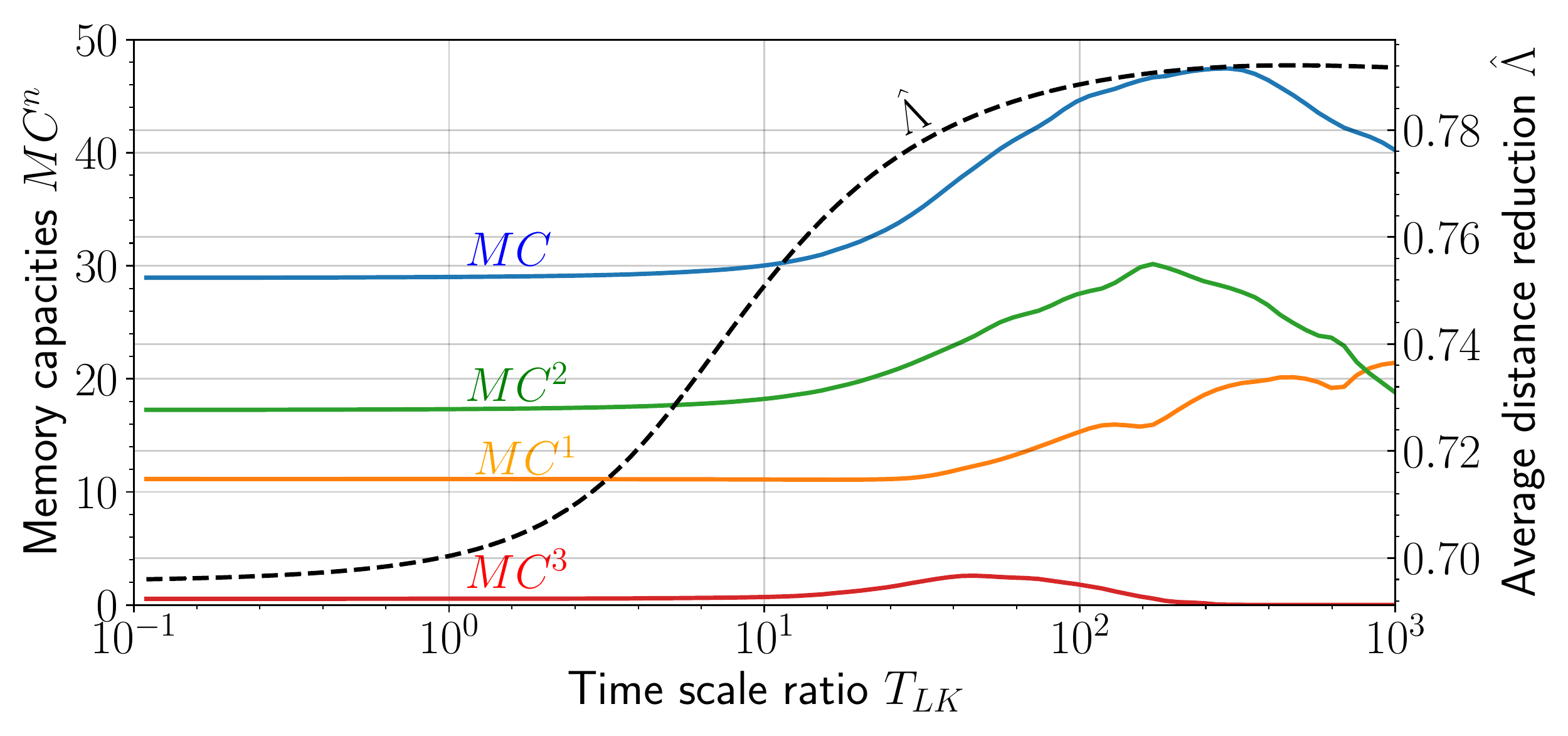}
    \caption{Linear, quadratic, cubic, and total memory capacities are shown as orange ,green, red, and blue lines as a function of the lifetime scale ratio $T_{LK}$. The sum of the distance reductions of the first 100 eigenvalues $\hat\Lambda= \frac{1}{N} \sum_{i=0}^{100} e^{\Re(\lambda_i)T}$ is plotted as a dashed black line. The increase of $\hat\Lambda$ coincides with the linear memory capacity.
    Parameters are $P=0.05$, $\kappa=0.1$, $T=410$, $\tau=500$, $N_{V}=50$.
    }%
    \label{fig:1d_delay_scan_sum_EW}%
\end{figure}%

As we have seen in Sec.~\ref{sec:2d} for a Laser with two dynamical degrees of freedom, the reservoir performance decreases when the real part of the eigenvalues becomes strongly negative. In such a case, the reservoir ''forgets'' the input too fast. 
Here we extend this idea to the case of the infinite-dimensional reservoir.

As long as the perturbation from the information fed into the system is small enough, one can think of all eigenvalues and their corresponding eigendirections as the available phase space of the reservoir computer. 
Thus, a higher phase space volume can lead to a more promising reservoir computer.
We introduce the average of the distance reduction $\Lambda$ by
\begin{align}
\label{eq:hatLambda}
    \hat{\Lambda} = \frac{1}{N}\sum_{k=0}^N e^{\Re(\lambda_k)T}.
\end{align}
It describes the average distance reduction of the $N$ slowest eigendirections.
Since only a finite number of complex eigenvalues lie to the right of a line parallel to the imaginary axis, all eigendirections except a finite number are strongly contracting, i.e., possess strongly negative real parts \cite{HAL93}.
This implies the possibility of considering only a finite number $N$ of eigenvalues in Eq. \eqref{eq:hatLambda}.

Figure \ref{fig:1d_delay_scan_sum_EW} depicts the memory capacities and the distance reduction $\hat\Lambda$ as a function of the timescale ratio $T_{LK}$, or in other words, the evolution of the memory capacities along the transition from a Class A to a Class B laser system with delayed feedback.
Similarly to the case without feedback in Fig.~\ref{fig:class_B_eigenvalue_memory_cap}, the memory capacity stays about constant for $T_{LK}\lesssim 2$, and increases when the additional dimensions become available by the reservoir for $T_{LK}\gtrsim 2$.
The increase of $\hat{\Lambda}$ 
coincides with the increase of the linear memory.
The higher orders show a similar trend, but are, in general, more involved and should be investigated more deeply.
Thus, the knowledge of the eigenvalues provides a qualitative prediction of the linear memory capacity. 

\begin{figure}%
    \centering
    \includegraphics[width=\textwidth]{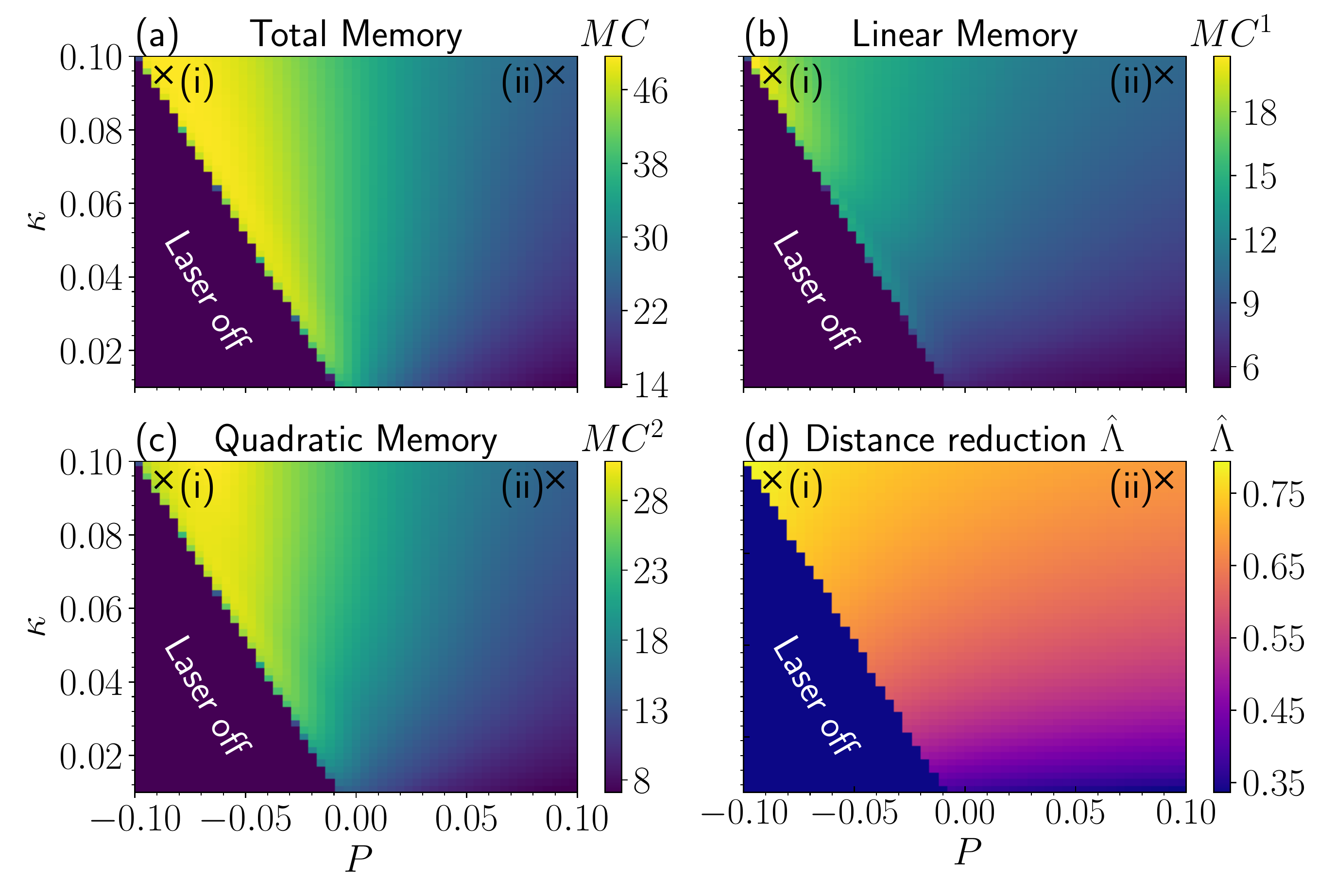}
    \caption{2-dimensional parameter scan in the plane of feedback strength $\kappa$ and pump $P$. Colorcoded are the linear (a), quadratic (b), total memory (c) capacities, and the (d) average distance reduction $\hat{\Lambda} = \frac{1}{N} \sum_{i=0}^{N} e^{\Re(\lambda_i)T}$ for the first 100 eigenvalues $N=100$. The two crosses indicate the parametervalues used in Fig. \ref{fig:1d_comparison_optimal_nonOptimal_point}. Parameters are $T_{LK}=1.0$, $T=350$ and $\tau=500$, $N_V=50$.}%
    \label{fig:2D-scan_delay_sum_EW}%
\end{figure}

To give a broader overview, we perform a 2-dimensional parameter scan along the feedback strength $\kappa$ and pump $P$ (shown in Fig. \ref{fig:2D-scan_delay_sum_EW}) and plot the linear, quadratic, and total memories as a color-code. Bright regions correspond to high memory capacities, and dark to low memory capacities.
Additionally, in Fig. \ref{fig:2D-scan_delay_sum_EW}(d), the sum of the average distance reduction $\hat{\Lambda}$ is color-coded within the same 2-dimensional parameter plane spanned by $\kappa$ and $P$.
Comparing the three memory capacity scans with $\hat{\Lambda}$, we can see a close relationship between them.
Thus, $\hat{\Lambda}$ is a very good indicator for choosing well-performing reservoir computers. This saves a lot of computational efforts, as the eigenvalues can be computed in a fraction of the time needed to compute the memory capacities.

\begin{figure}%
    \centering
    \includegraphics[width=0.90\textwidth]{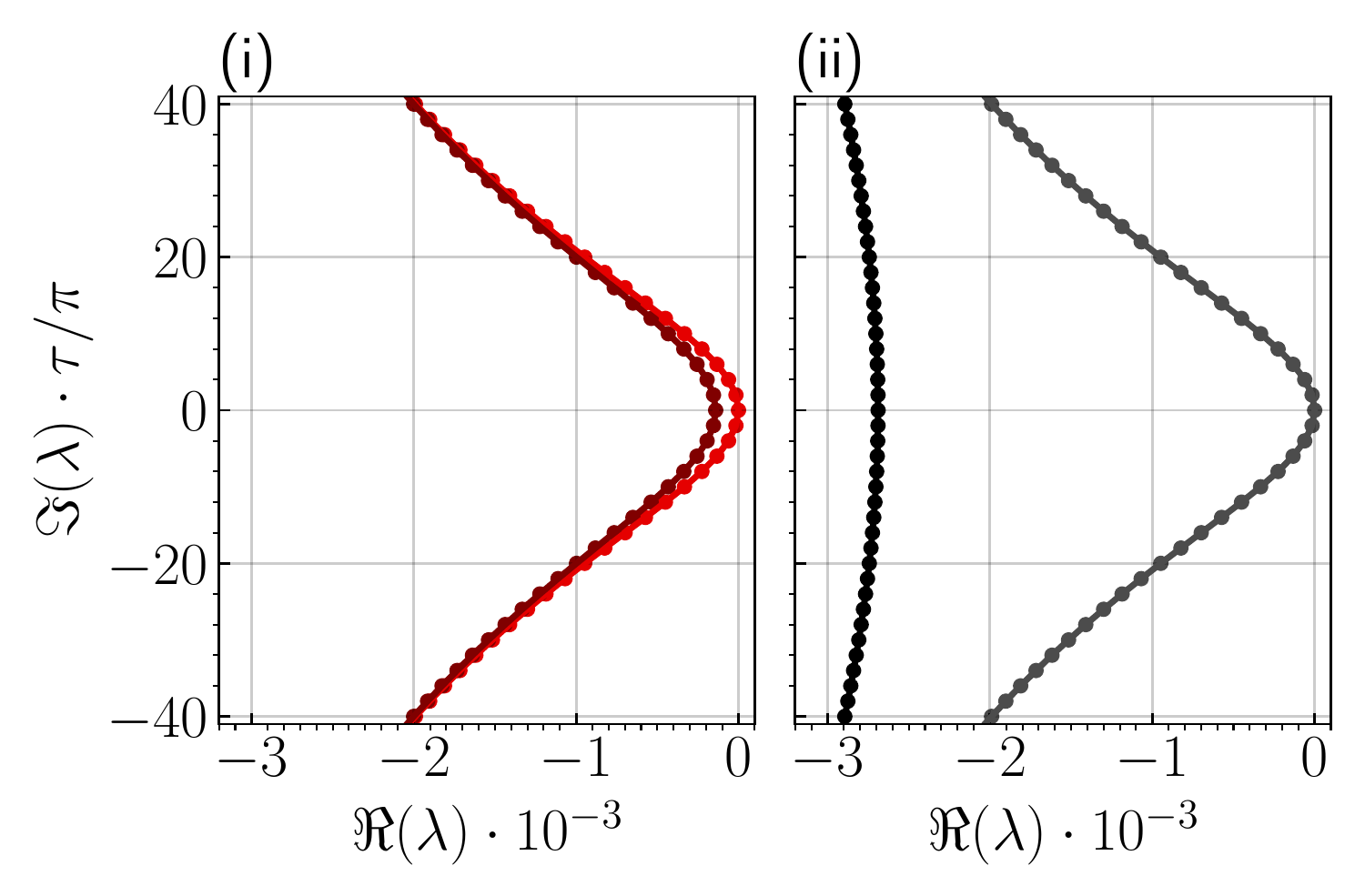}
    \caption{Pseudocontinous eigenvalue spectra for two different parameter sets given by Eq. \ref{eq:pcs1} (bright) and Eq. \ref{eq:pcs2} (dark) plotted for (i) $P=-0.095$ and (ii) $P=0.095$. $\kappa=0.1$, $\tau=500$, and $T_{LK} = 1.0$. The value of $T_{LK}=1.0$ corresponds to a solitary laser operating between Class A and Class B. For pump values slightly above threshold $P_{th}=-\kappa$ (i), the eigenvalue spectrum has more eigenvalues with real parts close to $0$.
    \label{fig:EW_spectrum}%
    }%
\end{figure}

To illustrate possible configurations of eigenvalues and their connection to $\hat{\Lambda}$, 
we chose two different parameter setups 
in  Fig.~\ref{fig:EW_spectrum}: $P=-0.095$   (a) and  $P=0.095$ (b), with the other parameters fixed.
The two parameter setups are marked as black crosses in Fig.~\ref{fig:2D-scan_delay_sum_EW}, which correspond to parameters close to and well above threshold. 
The first parameter set (Fig.~\ref{fig:EW_spectrum}(i)) corresponds to an eigenvalue spectrum for an operation point close above the threshold with a low power output. 
Here the laser system possesses more eigenvalues with real parts close to 0, thus it has many slowly contracting eigendirections which means $\hat{\Lambda}$ is closer to 1.
Calculating $\hat{\Lambda}$ for the first parameter set yields $\hat{\Lambda} = 0.75$.
The second parameter set (Fig.~\ref{fig:EW_spectrum}(ii)) corresponds to a laser that is operated high above threshold. This laser has less slowly contracting eigendirections, i.e. $\hat{\Lambda}$ is closer to 0.
Calculating $\hat{\Lambda}$ for the second parameter set yields $\hat{\Lambda} = 0.6$.

\begin{figure}%
    \centering
    \includegraphics[width=\textwidth]{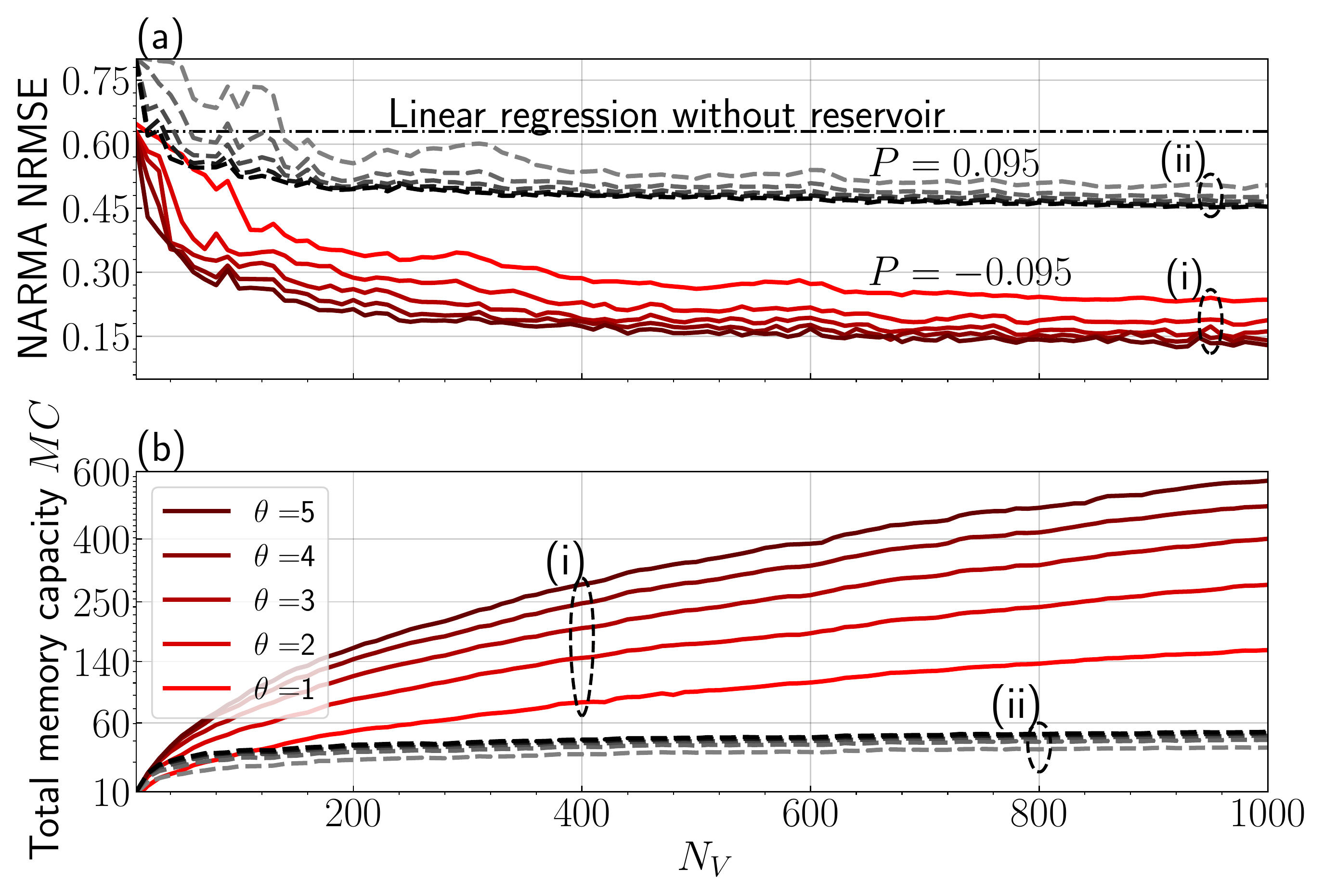}
    \caption{Computation error NARMA10 NRMSE (a) and the total memory capacity (b) for the two parameter values:
    $P=0.095$ (dashed lines) and 
    $P=-0.095$ (solid lines), see also  
    crosses in Fig.~\ref{fig:2D-scan_delay_sum_EW}. 
    Other parameters: $\tau = 1.41T$, $\kappa=0.1$, and $T_{LK}=1.0$. 
    On the $x$ axis the number of virtual nodes $N_V$ is shown.
    Different brightness correspond to different distances $\theta$ between the virtual nodes. 
    The dashed black line shows the minimum reached with a linear regression without a reservoir.
    }%
    \label{fig:1d_comparison_optimal_nonOptimal_point}%
\end{figure}

Now we use the insights gained from the distance reduction $\Lambda$ and from the angular distance $\Phi$ and test the reservoir computer performances for the two parameter sets from Fig.~\ref{fig:EW_spectrum}(i) and Fig.~\ref{fig:EW_spectrum}(ii), marked as black crosses in Fig.~\ref{fig:2D-scan_delay_sum_EW}.
The performance is quantified by evaluating both the memory capacity and the prediction error (NRMSE) for the NARMA10 task shown in Fig.~\ref{fig:1d_comparison_optimal_nonOptimal_point}(b) and Fig.~\ref{fig:1d_comparison_optimal_nonOptimal_point}(a).

On the horizontal axis, we change the number of virtual nodes $N_V$, i.e., we increase the number of readout dimensions, which should, naively thinking, increase the performance of the reservoir.
While we do this, we keep the distance between the virtual nodes $\theta$ the same for 5 different cases of $\theta$ from $\theta=1$ up to $\theta=5$, shown in black and red with decreasing brightness for the optimized (Fig.~\ref{fig:EW_spectrum}(i)) and not-optimized (Fig.~\ref{fig:EW_spectrum}(ii)) point respectively.
Increasing the virtual node distance $\theta$ should reduce the linear dependency of the nodes, as the time between two responses is increased.
This is obviously dependent on the reaction time of the system, which is also given by the eigenvalues of the system.
Thus the influence of increasing $\theta$ on the slowly contracting eigendirections (Fig.~\ref{fig:EW_spectrum}(i)) is pronounced compared to the one with fast eigendirection (Fig.~\ref{fig:EW_spectrum}(ii)).
The increase of $\theta$ also effectively increases the clock cycle $T=N_V \theta$ and thus the delay time $\tau= 1.41 T$.
We want to emphasize, that this does not alter the general trend of $P=-0.095$ (Fig.~\ref{fig:EW_spectrum}(i)) having many slowly contracting eigendirections compared to $P=0.095$ (Fig.~\ref{fig:EW_spectrum}(ii)).

The results indicate that even though the number of virtual nodes increases, the NARMA10 error for the case with the low distance reduction $\hat{\Lambda}$ (ii) does not go below $0.45$.
On the other hand the case with the high $\hat{\Lambda}$, our optimal case (i), reaches very small errors below $0.15$, a factor 3 better than the low distance reduction $\hat{\Lambda}$ (ii) case. We also want to emphasize that the simulation was done for a high noise value of $D_{noise}=10^{-7}$. Simulating the system without any noise $D_{noise}=0$ NARMA10 errors (NRMSE) of down to $0.05$ were reached.
We conclude that a high distance reduction is very beneficial for the performance.

As a dashed black line in Fig.~\ref{fig:1d_comparison_optimal_nonOptimal_point}(a) we additionally show the minimum reached by a linear regression without a reservoir.
Every reservoir setup with results below this line has higher performance and thus can be considered an improvement on the NARMA10 task.
We included this here to emphasize the reduction of the NARMA10 error (NRMSE) by the eigenvalue analysis, for which an improvement of about a factor 4 is reached to the linear regression without reservoir.

The total memory capacities for the two cases (i) and (ii) are shown in Fig.~\ref{fig:1d_comparison_optimal_nonOptimal_point}(b).
We can see the same trend: the memory capacity reaches a limit for the case with low $\hat{\Lambda}$, whereas the improved case with the highest $\hat{\Lambda}$ increases in its memory capacity further for higher $N_V$.
The results suggest that operation points with high distance reduction $\hat{\Lambda}$ (solid lines in Fig.~\ref{fig:1d_comparison_optimal_nonOptimal_point}) pinpoint to well-performing reservoir computers.

\section{Conclusion}
We have shown that the eigenvalue spectrum analysis of a dynamical system used as a reservoir (e.g. a laser described by the Lang-Kobayashi system) is capable of predicting good reservoir computing operation points. 
Because of the available analytical and numerical tools for the description of the eigenvalue spectrum, such analysis can be readily applied for different dynamical systems, which are used as reservoirs with operating points close to an equilibrium.
The eigenvalue method is of magnitudes faster to compute and could help in numerically predicting good reservoir computers for experimental setups.

Due to the relation between the eigenvalue spectrum and the performance of the delay-based reservoir computing, 
the central message of this paper is twofold:
First, the eigenvalues must be off-resonant, where the
resonance condition is given in terms of the imaginary parts
of the eigenvalues. 
Namely, the product of the latter with the input clock-cycle should be away from values of $k\pi$. Importantly, such resonances appear for all critical eigenvalues at almost the same parameter values, due to general properties of the spectrum of delay systems with large delays \cite{LIC11}. Therefore, such an off-resonant condition plays an important role even when the reservoir's effective dimensionality is high.

The second conclusion is that, for optimal performance, the spectrum must be close to criticality. This closeness is measured by the real part of the eigenvalue spectrum, which should be close to zero and negative. In this paper, we propose the average distance reduction as a measure $\hat \Lambda$ for such closeness, which is  given by Eq.~\eqref{eq:hatLambda}.

The presented timescale analysis, i.e., eigenvalue analysis of the reservoir, has some further advantages. 
The eigenvalues do not only yield the timescale on which the system forgets, but also the timescale on which the system learns (for small inputs). 
Thus, through the eigenvalues one can construct a system which is either fast and has short term memory or a system similar to what we did in this work, which is slow and has more memory capacity further into the past. 
We would like to use this in future works in constructing reservoirs capable of tackling many different tasks.

\appendix

\section{Memory capacity expression \eqref{eq:mpsi_mem_capacity} \label{sec:APP}}
We show how the expression for the memory capacity from 
\cite{DAM12} can be rewritten in the form of Eq.~\eqref{eq:mpsi_mem_capacity}. 
From Dambre. et. al. \cite{DAM12},  the capacity to approximate a target data $\mathbf{\hat{y}}$ is given by
\begin{align}
    C_L[\mathsf{S},\mathbf{\hat{y}}] = \frac{\sum_{ij} \langle \hat{y} s_i \rangle_L \langle s_i s_j \rangle_L^{-1} \langle s_j \hat{y} \rangle_L}{\langle \hat{y}^2 \rangle_L}.
\end{align}
Here $s_i$ is the i-th readout of the $M$ system responses for the $l$-th input-target pair, $\langle \nu \rangle_L = \frac{1}{L}\sum_{l=1}^L \nu_l$ is the average over all input-output paris $L$ and $\langle s_i s_j \rangle_L^{-1}$ is the inverse of $\langle s_i s_j \rangle_L$.
One can insert the average over all input-output pairs yielding
\begin{align}
    C_L[\mathsf{S},\mathbf{\hat{y}}] = \frac{\sum_{ij} \sum_{l=1}^L \hat{y}_l s_{i,l} (\sum_{l=1}^L s_{i,l}s_{j,l})^{-1} \sum_{l=1}^L s_{j,l} \hat{y}_l }{\sum_{l=1}^L \hat{y}_l^2 }.
\end{align}
In the denominator $\sum_{l=1}^L \hat{y}_l^2$ can be substituted with the square norm of the target vector $\sum_{l=1}^L \hat{y}_l^2 = \norm{\mathbf{\hat{y}}}^2$.
The first term $\hat{y}_l s_{i,l}$ is the $i$-th system response to the $l$-th input-output pair ($i$-th column and $l$-th row) multiplied with the $l$-th target. Summing over all input-output pairs $L$, this is the same as the $i$-th entry of the matrix product
\begin{align}
    \sum_{l=1}^L \hat{y}_l s_{i,l} = (\mathbf{\hat{y}}^T\mathsf{S})_i
\end{align}
The same reasoning applies to $(\sum_{l=1}^L s_{i,l}s_{j,l})$ yielding in
\begin{align}
    \left(\sum_{l=1}^L s_{i,l}s_{j,l}\right)^{-1} = (\mathsf{S}^T\mathsf{S})^{-1}_{i,j}
\end{align}
$\sum_{l=1}^L s_{j,l} \hat{y}_l$ is just the transposed case of $\sum_{L=1}^L \hat{y}_l s_{i,l}$, thus $\sum_{l=1}^L s_{j,l} \hat{y}_l=(\mathsf{S}^T\mathbf{\hat{y}})_j$.
Summing over all responses $i$ and $j$ is equivalent to the matrix product
\begin{align}
    \sum_{ij} (\mathbf{\hat{y}}^T\mathsf{S})_i (\mathsf{S}^T\mathsf{S})^{-1}_{i,j} (\mathsf{S}^T\mathbf{\hat{y}})_j = \mathbf{\hat{y}}^T \mathsf{S} (\mathsf{S}^T\mathsf{S})^{-1} \mathsf{S}^T \mathbf{\hat{y}},
\end{align}
with which we have reached Eq. \eqref{eq:mpsi_mem_capacity}
\begin{align} 
    \text{C} = \frac{\mathbf{\hat{y}}^T \mathsf{S} (\mathsf{S}^T\mathsf{S})^{-1} \mathsf{S}^T \mathbf{\hat{y}}}{\norm{\mathbf{\hat{y}}}^2}.
\end{align}

\section*{Acknowledgment}
The authors thank Florian Stelzer and Mirko Goldmann for fruitfull discussions.
This study was funded by the "Deutsche Forschungsgemeinschaft" (DFG) in the framework of SFB910 and project 411803875.

\bibliographystyle{ieeetr}
%
\bibliography{ref.bib}

\end{document}